\documentclass[sigconf,natbib=false]{acmart}

\usepackage{balance}

\usepackage{amsmath}
\usepackage{graphicx}
\usepackage{hyperref}
\usepackage[linesnumbered,ruled]{algorithm2e}
\usepackage[backend=bibtex,doi=false,isbn=false,url=false,maxbibnames=1,style=trad-abbrv]{biblatex}
\usepackage{xcolor}
\usepackage{soul}
\usepackage{hhline}
\usepackage[normalem]{ulem}
\usepackage{stmaryrd}

 \usepackage{multirow}
 \usepackage{multicol}
 \usepackage{booktabs}
 \usepackage{fancyhdr}
 \usepackage{tabularx}
 \usepackage{float}
\usepackage{comment}
\usepackage{arydshln}

\AtBeginDocument{%
  \providecommand\BibTeX{{%
    \normalfont B\kern-0.5em{\scshape i\kern-0.25em b}\kern-0.8em\TeX}}}

\begin{document}

\title{Improving Graph Neural Networks with Simple Architecture Design}

\author{Sunil Kumar Maurya}
\email{skmaurya@net.c.titech.ac.jp}
\orcid{}
\affiliation{%
  \institution{Tokyo Institute of Technology}
  \streetaddress{}
  \city{Tokyo}
  \state{}
  \country{Japan}
  \postcode{}}
\author{Xin Liu}
\email{xin.liu@aist.go.jp}
\affiliation{%
  \institution{AIRC, AIST}
  \streetaddress{}
  \city{Tokyo}
  \state{}
  \country{Japan}
  \postcode{}
}
\author{Tsuyoshi Murata}
\email{murata@c.titech.ac.jp}
\affiliation{%
  \institution{Tokyo Institute of Technology}
  \streetaddress{}
  \city{Tokyo}
  \state{}
  \country{Japan}
  \postcode{}
}


\renewcommand{\shortauthors}{Maurya et al.}

\begin{abstract}
  Graph Neural Networks have emerged as a useful tool to learn on the data by applying additional constraints based on the graph structure. These graphs are often created with assumed intrinsic relations between the entities. In recent years, there have been tremendous improvements in the architecture design, pushing the performance up in various prediction tasks. In general, these neural architectures combine layer depth and node feature aggregation steps. This makes it challenging to analyze the importance of features at various hops and the expressiveness of the neural network layers. As different graph datasets show varying levels of homophily and heterophily in features and class label distribution, it becomes essential to understand which features are important for the prediction tasks without any prior information. In this work, we decouple the node feature aggregation step and depth of graph neural network and introduce several key design strategies for graph neural networks. More specifically, we propose to use softmax as a regularizer and "Soft-Selector" of features aggregated from neighbors at different hop distances; and "Hop-Normalization" over GNN layers. Combining these techniques, we present a simple and shallow model, Feature Selection Graph Neural Network (FSGNN), and show empirically that the proposed model outperforms other state of the art GNN models and achieves up to 64\% improvements in accuracy on node classification tasks. Moreover, analyzing the learned soft-selection parameters of the model provides a simple way to study the importance of features in the prediction tasks. Finally, we demonstrate with experiments that the model is scalable for large graphs with millions of nodes and billions of edges.
  
  Source code at https://github.com/sunilkmaurya/FSGNN
\end{abstract}





\maketitle

\section{Introduction}
\label{introduction}
Graph Neural Networks (GNNs) have opened a unique path to learning on data by leveraging the intrinsic relations between entities that can be structured as a graph. By imposing these structural constraints, additional information can be learned and used for many types of prediction tasks. With rapid development of the field and easy accessibility of computation and data, GNNs have been used to solve a variety of problems like node classification \cite{kipf_semi-supervised_2017,velickovic_graph_2017,abu-el-haija_mixhop_2019,chen_simple_2020}, link prediction \cite{ying_graph_2018,berg_graph_2017,chami_hyperbolic_2019}, graph classification \cite{ying_hierarchical_2018,zhang_end--end_2018}, prediction of molecular properties \cite{gilmer_neural_2017,madhawa_graphnvp_2019}, natural language processing \cite{marcheggiani_encoding_2017}, node ranking \cite{maurya_fast_2019} and so on.

In this work, we focus on the node classification task using graph neural networks. Since the success of early GNN models like GCN \cite{kipf_semi-supervised_2017}, researchers have successively proposed numerous variants \cite{wu_comprehensive_2019} to address various shortcomings in model training and to improve the prediction capabilities. Some of the techniques used in these variants include  neighbor sampling \cite{hamilton_inductive_2017,chen_fastgcn:_2018}, attention mechanism to assign different weights to neighbors \cite{velickovic_graph_2017}, use of Personalized PageRank matrix instead of adjacency matrix \cite{klicpera_predict_2018} and simplified model design \cite{wu_simplifying_2019}. Also, there has been a growing interest in making the models deeper by stacking more layers and using the residual connections to improve the expressiveness of the model \cite{rong_dropedge_2020,chen_simple_2020}.
However, most of these models by design are more suitable for homophily datasets where nodes linked to each other are more likely to belong in the same class. They may not perform well with heterophily datasets which are more likely to have nodes with different labels connected together. Zhu et al. \cite{zhu_beyond_2020} highlight this problem and propose node's ego-embedding and neighbor-embedding separation to improve performance on heterophily datasets.

In general, GNN models combine feature aggregation and transformation using a learnable weight matrix in the same layer, often referred to as graph convolutional layer.
These layers are stacked together with the non-linear transformation (e.g., ReLU) and regularization(e.g., Dropout) as a learning framework on the graph data. Stacking the layers also has the effect of introducing powers of adjacency matrix (or laplacian matrix), which helps to generate a new set of features for a node by aggregating neighbor's features at multiple hops, thus encoding the neighborhood information. The number of these unique features depends on the propagation steps or the depth of the model. The final node embeddings are the output of just stacked layers or, for some models, also has skip connection or residual connection combined at final layer.

However, such a combination muddles the distinction between the importance of features and expressiveness of MLP. It becomes challenging to analyze which features are essential and how much expressiveness MLP requires over a specific task. To overcome this challenge, we provide a framework to treat the feature propagation and learning separately. With this freedom, we propose a simple GNN model with three unique design considerations: Soft-selection of features using softmax function, Hop-Normalization, and unique mapping of features. With experimental results, we show that our simple 2-layer GNN outperforms other state-of-art GNN models (both shallow and deep) and achieves up to 64\% higher
node classification accuracy. In addition, analyzing the model parameters gives us an insight into identifying which features are most responsible
for classification accuracy. One interesting observation we find is regarding Chameleon and Squirrel datasets. These are dense graph datasets and are generally regarded as being low-quality heterophily datasets. However, in our experiments with our proposed model, we find them to be showing strong heterophily properties with improved classification results. 

Furthermore, we demonstrate that due to the simple design of our model, it can scale up for very large graph datasets. We run experiments on ogbn-papers100M dataset, which is the largest publicly available node classification dataset, and achieve higher accuracy than the state of the art models.

The rest of the paper is organized as follows: Section \ref{preliminaries} outlines formulation of graph neural networks and details node classification task. In Section \ref{propose_arch}, we discuss design strategies for GNNs and propose the GNN model FSGNN. In Section \ref{related_work}, we briefly introduce relevant GNN literature. Section \ref{experiments} contains the experimental details and comparison with other GNN models. In Section \ref{discussion}, we empirically analyze our proposed design strategies and their effect on the model's performance. Section \ref{conclusion} summarizes the paper.

\section{Preliminaries}
\label{preliminaries}

Let $G = (V,E)$ be an undirected graph with $n$ nodes and $m$ edges. For numerical calculations, graph is represented as adjacency matrix denoted by $A\in \{0,1\}^{n\times n}$ with each element $A_{ij}=1$ if there exists an edge between node $v_i$ and $v_j$, otherwise $A_{ij}=0$. If self-loops are added to the graph then, resultant adajcency matrix is denoted as $\Tilde{A} = A+I$. Diagonal degree matrix of $A$ and $\Tilde{A}$ are denoted as $D$ and $\Tilde{D}$. Each node is associated with a d-dimensional feature vector and the feature matrix for all nodes is represented as $X \in \mathbb{R} ^{n \times d}$.

\subsection{Graph Neural Networks}

Graph Neural Networks (GNNs) leverage feature propagation mechanism \cite{gilmer_neural_2017} to aggregate neighborhood information of a node and use non-linear transformation with trainable weight matrix to get the final embeddings for the nodes. Conventionally, a simple GNN layer is defined as 
\begin{equation}
  \label{eq:homophily_gnn}
  H^{(i+1)} = \sigma (\Tilde{A}_{sym}H^{(i)}W^{(i)})
\end{equation}

where $\Tilde{A}_{sym} = \Tilde{D}^{-\frac{1}{2}} \Tilde{A}\Tilde{D}^{-\frac{1}{2}}$ is a symmetric normalized adjacency matrix with added self-loops. $H^i$ represents features from the previous layer, $W^i$ denotes the learnable weight matrix, and $\sigma$ is a non-linear activation function, which is usually ReLU in most implementation of GNNs. However, this formulation is suitable for homophily datasets as features are cumulatively aggregated i.e. node's own features are added together with neighbor's features. For heterophily datasets, we require a propagation scheme to separate features of neighbors from node's own features. So we use the following formulation for the GNN layer,

\begin{equation}
      \label{eq:heterophily_gnn}
      H^{(i+1)} = \sigma (A_{sym}H^{(i)}W^{(i)})
\end{equation}
where $A_{sym} = D^{-\frac{1}{2}} AD^{-\frac{1}{2}}$ is symmetric normalized adjacency matrix without added self-loops. To combine features from multiple hops, concatenation operator can be used before the final layer.

Following the conventional GNN formulation using $\Tilde{A}$, a simple 2-layered GNN can be represented as \cite{kipf_semi-supervised_2017},

\begin{equation}
    \label{eq:gnn_2layer}
    Z = \Tilde{A}_{sym}\sigma(\Tilde{A}_{sym}XW^{(0)})W^{(1)}
\end{equation}

\subsection{Node Classification}

Node classification is an extensively studied graph based semi-supervised learning problem. It encompasses training the GNN to predict labels of nodes based on the features and neighborhood structure of the nodes. GNN model is considered as a function $f(X,A)$ conditioned on node features $X$ and adjacency matrix $A$. Taking the example of Eq. \ref{eq:gnn_2layer}, GNN aggregates the features of two hops of neighbors and outputs $Z$. Softmax function is applied row-wise, and cross-entropy error is calculated over all labeled training examples. The gradients of loss are back-propagated through the GNN layers. Once trained, the model can be used for the prediction of labels of nodes in the test set.

\begin{figure*}[h]
    \centering
    \includegraphics[width=0.9\textwidth]{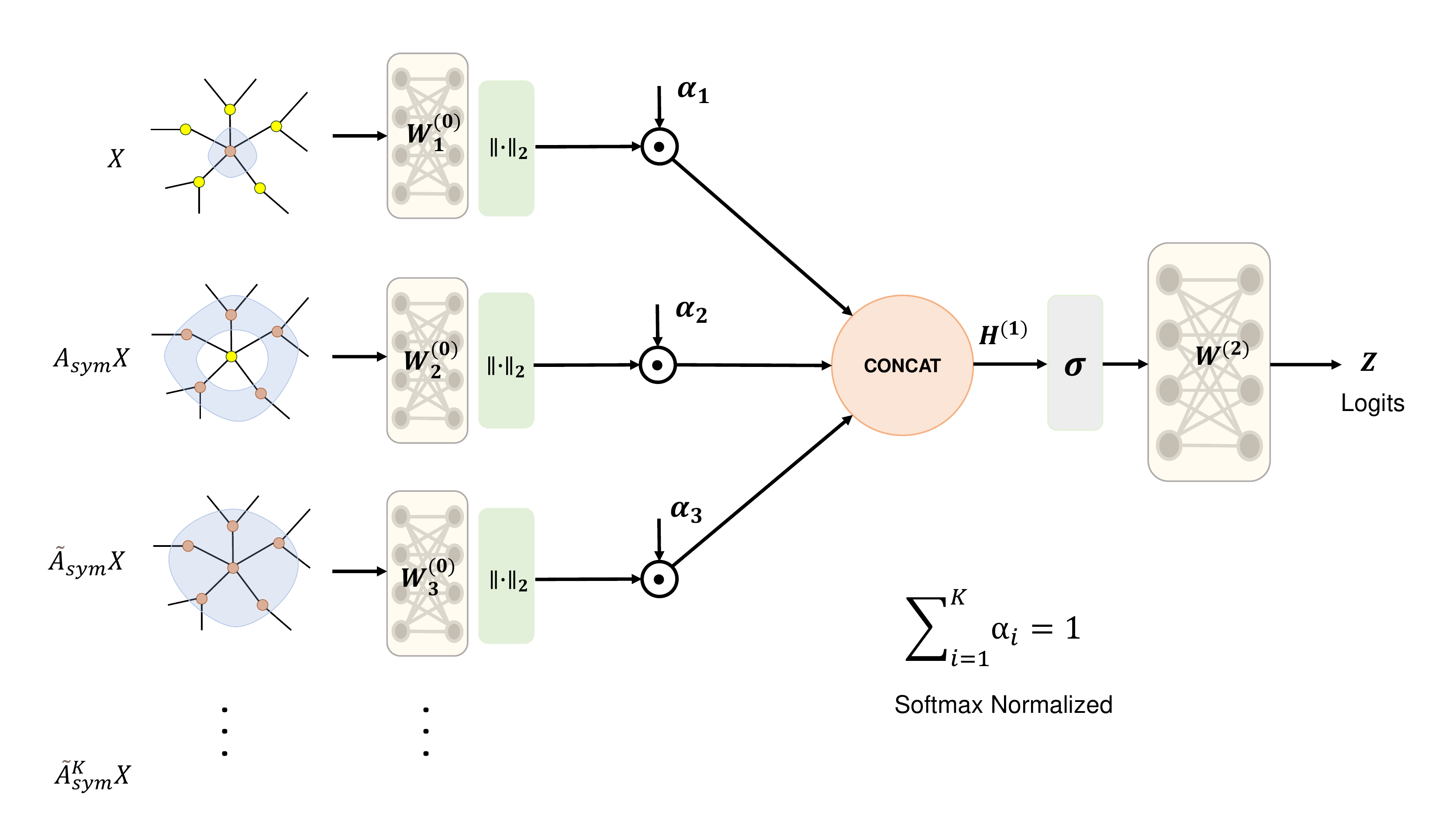}
    \caption{Figure shows model diagram of FSGNN. Input features are generated based on powers of $A$ and $\Tilde{A}$.}
    \label{fig:model_diagram}
\end{figure*}

\subsection{Homophily vs Heterophily}
Node classification problem relies on the graph structure and features of the nodes to identify the labels of the node. Under homophily, nodes are assumed to have neighbors with similar features and labels. Thus, the cumulative aggregation of node's self-features with that of neighbors reinforce the signal corresponding to the label and help to improve accuracy of the predictions. While in the case of heterophily, nodes are assumed to have dissimilar features and labels. In this case, the cumulative aggregation will reduce the signal and add more noise causing neural network to learn poorly and causing drop in performance. Thus it is essential to have node's self-features separate from the neighbor's features. In real-world datasets, homophily and heterophily levels may vary, hence it is optimal to have both aggregation schemes (Eq. \ref{eq:homophily_gnn} \& \ref{eq:heterophily_gnn})

\section{Proposed Architecture}
\label{propose_arch}
For the design of a GNN with good generalization capability and performance, there are many aspects of the data that needs to be considered. The feature propagation and aggregation scheme is governed by if the class label distribution has strong homophily or heterophily or some combination of both. The number of hops (and depth of the model for many GNN models) for feature aggregation are dependent on graph structure and size as well as label distribution among neighbors of the nodes. Also, the type and amount of regularization during training needs to be decided, for example, using dropout on input features or on graph edges.

Keeping these aspects under consideration, we propose three design strategies that help to create a versatile and simple GNN model.

\vspace{5mm}
\subsection{Design Strategies for GNNs}
\subsubsection{Decouple feature generation and representation learning}\hfill

As discussed in Sec. 2.1, these features can be aggregated cumulatively (homophily-based) or non-cumulatively (heterophily-based). Moreover, the features can also be combined based on some arbitrary criteria. We assume a function, 
$$g(X,A,K) \mapsto \{X_1,X_2, \dotsc ,X_p\}$$
The function takes $X$ as node features matrix, $A$ as an adjacency matrix, $K$ as the power of the adjacency matrix or number of hops to propagate features and outputs a set of aggregated features. These features then can be combined using sum or concatenation operation to get final representation of the node.  However, in the node classification task, for a given label distribution, only a subset of these features are useful to predict the label of the node. For example, features of node's neighbors that lie at a greater distance in the graph may not be sufficiently informative or useful for node's label prediction.

Conventionally, GNN models have feature propagation and transformation combined into a single layer, and the layers are stacked together. This step makes it difficult to distinguish the importance of the features and the role of MLP. To overcome this limitation, we propose to separate the feature generation step and representation learning over features separately. This provides us with three main benefits. 

\renewcommand{\labelenumi}{(\roman{enumi})}
\begin{enumerate}
    \item Features generated for nodes are not constrained by the design of the GNN model. We get the freedom to choose the feature set as required by the problem and the neural network design, which is sufficiently expressive.
    \item We can precompute and fix the node features set and experiment with the neural network architectures for the best performance. Precomputing features also helps to scale the training of the model for large graphs with batchwise training.
    \item In conventional GNN setting, stacking many layers also causes oversmoothing of node features \cite{chen_measuring_2019} and adversely affects the performance of the model. Recently proposed models use skip connection or residual connection to overcome this issue. However, they fail to demonstrate which features are useful. We provide an alternate scheme where the model can learn weights that identify which features are useful for the prediction task.
\end{enumerate}

For the model design, instead of a single input channel, we propose to have all these features as input in parallel. Please refer Fig.\ref{fig:model_diagram} for the illustration.  Each feature is mapped to a separate linear layer. Hence the linear transformations are uniquely learned for all input features. 

\subsubsection{Feature Selection}\hfill

As features are aggregated over many hops, some features are useful and correlate with the label distribution, while others are not very useful for learning and act more like the noise for the model. As we propose to input the feature set in parallel channels, we can design the model to learn which features are more relevant for lower loss value and giving higher weights to those features while simultaneously reducing the weights on other features. We propose to weight these features with a single scalar value that is multiplied to each input feature matrix and impose a constraint on these values by softmax function. Let $\alpha_i$ be the scalar value for the $i^{th}$ feature matrix, then $\alpha_i$ scales the magnitude of the features as $\alpha_i X_iW^{(0)}_i$. Softmax function is used in deep learning as a non-linear normalizer, and its output is often practically interpreted as probabilities. Before training, the scalar values corresponding to each feature matrix are initialized with equal values and softmax is applied on these values. The resultant normalized values $\alpha_i$ are then multiplied with the input features, and the concatenation operator is applied. Considering $L$ number of input feature matrices $X_l,\: l\in\{1\:..\:L\}$ , the formulation can be described as,

\begin{equation}
    H^{(1)} = \bigparallel^{L}_{l=1} \alpha_lX_lW_{l}^{(0)}
\end{equation}
$$ \textrm{where         }   \sum_{l=1}^{L}\alpha_{l} = 1$$

 While training, the scalar values of relevant features corresponding to the labels increase towards 1 while others decrease towards 0. The features that are not useful and represent more noise than signal have their magnitudes reduced with corresponding decreasing in their scalar values. Since we are not using a binary selection of features, we term this selection procedure as "soft-selection" of features.

This formulation can be understood in two ways. As GNNs have represented with a polynomial filter, 

\begin{equation}
    g_\theta(P) = \sum_{k=0}^{K-1}\theta_kP^k
\end{equation}

where $\theta \in \mathbb{R}^K$ is a vector of polynomial coefficients and P can be adjacency matrix \cite{kipf_semi-supervised_2017}\cite{chen_simple_2020}, laplacian matrix \cite{nt_stacked_2020} or PageRank based matrix \cite{berberidis_adaptive_2019}. As the polynomial coefficients are scalar parameters then our scheme can be considered as applying regularization on these parameters using the softmax function. The other way to look is to simply consider it as a weighting scheme. As the input features can be arbitrarily chosen, and instead of a scalar weighting scheme, a more sophisticated scheme can be used. 

 For practical implementation, since all weights are initialized as equal, they can be set equal to 1. After normalizing with softmax function, the individual scalar values becomes equal to $1/L$. During training, these values change, denoting the importance of the features. In some cases, initial $\alpha_l = 1/L$ value may be too small and may adversely affect training. In that case, a constant $\gamma$ may be multiplied after softmax normalization to increase the initial magnitude as $\gamma\alpha_lX_lW_{l}^{(0)}$. Since $\gamma$ remains constant during the training, it does not affect the softmax regularization of the scalar parameters. 
 
 As the scalar values affect the magnitude of the features, they also affect the gradients propagated back to the linear layer, which transforms the input features. Hence it is important to have a unique weight matrix for each input feature matrix.

\subsubsection{Hop-Normalization}\hfill

The third strategy we propose is Hop-Normalization. It is a common practice in the deep learning field to use different types of normalization schemes, for example, batch normalization \cite{ioffe_batch_2015}, layer normalization, weight normalization, and so on. However, in graph neural network frameworks, normalization of activations after hidden layers are not commonly used. It may be in part due to the common practice of normalizing node/edge features and symmetric/non-symmetric normalization of the adjacency matrix.

We propose to normalize all aggregated features from different hops after linear transformation, hence the term "Hop-Normalization". We propose row-wise L2-normalize the hidden layer activations as,

\begin{equation}
    h_{ij} = \frac{h_{ij}}{\parallel h_{i} \parallel_2}
\end{equation}

 where $h_{i}$ represents the $i^{th}$ row vector of activations and $h_{ij}$ represents individual values. L2-normalization scales the node embedding vectors to lie on the "unit sphere". In the later section, we empirically show significant improvements in the performance of the model with the use of this scheme.

\subsection{Feature Selection Graph Neural Network}

Combining the design strategies proposed earlier, we propose a simple and shallow (2-layered) graph GNN model called Feature Selection Graph Neural Network (FSGNN). Figure \ref{fig:model_diagram} shows the diagrammatic representation of our model. Input features are precomputed using $A_{sym}$ and $\Tilde{A}_{sym}$ and transformed using a linear layer unique to each feature matrix. Hop-normalization is applied on the output activations of the first layer and weighted with scalar weights regularized by the softmax function. Output features are then concatenated and non-linearly transformed using ReLU and mapped to the second linear layer. Cross-entropy loss is calculated with output logits of second layer.

\setlength{\textfloatsep}{1mm}
\begin{algorithm}[h!]
\DontPrintSemicolon
\caption{Pseudo Code FSGNN (Forward propagation)}
\label{alg:fsgnn}
    \SetKwInOut{Input}{Input}\SetKwInOut{Output}{Output}
    \Input{~ $A_{sym}$; $\Tilde{A}_{sym}$; No. of hops $K$; weight matrices $W^{(k)}$; $\alpha$ vector of dimension 2K+1;   }
    \Output{~ Logits}
    \BlankLine
    $\alpha_i\leftarrow1.0,  i=1...2K+1$ \;
    $\alpha \leftarrow SOFTMAX(\alpha)$ \;
    $list\_mat \leftarrow [X]$ \;
    $X_A\leftarrow X$ \;
    $X_{\Tilde{A}} \leftarrow X$ \;

    \For{$k=1...K$}{
        $X_A \leftarrow A_{sym}X_A$ \;
        $X_{\Tilde{A}} \leftarrow \Tilde{A}_{sym}X_{\Tilde{A}}$ \;
        $list\_mat.APPEND(\:X_A\:)$ \;
        $list\_mat.APPEND(\:X_{\Tilde{A}}\:)$ \;
        }
    $list\_cat = LIST()    $\;
    \For{$j=1...2K+1$}{
        $X_f \leftarrow list\_mat[j]$ \;
        $Out \leftarrow HOPNORM(\:X_fW_{j}^{(0)}\:) $ \;
        $list\_cat.APPEND(\:\alpha_j \odot Out\:)$ \;
    }
    $H^{(1)} \leftarrow CONCAT(\:list\_cat\:)$ \;
    $Z \leftarrow ReLU(\: H^{(1)}\:)W^{(2)}$

\end{algorithm}

\section{Related Work}
\label{related_work}

GNNs have emerged as an indispensable tool to learn graph-centric data. Many prediction tasks like node classification, link prediction, graph classification, etc. \cite{defferrard_convolutional_2016}\cite{kipf_semi-supervised_2017} introduced a simple end-to-end training framework using approximations of spectral graph convolutions. Since then, there has been a focus in the research community to improve the performance of GNNs, and a variety of techniques have been introduced. Earlier GNN frameworks utilized a fixed propagation scheme along all edges, which is sometimes not scalable for larger graphs. GraphSAGE\cite{hamilton_inductive_2017} and FastGCN\cite{chen_fastgcn:_2018} introduce neighbor sampling approaches in graph neural networks. GAT \cite{velickovic_graph_2017} introduces the use of the attention mechanism to provide weights to features that are aggregated from the neighbors. APPNP \cite{klicpera_predict_2018}, JK \cite{xu_representation_2018} and Geom-GCN \cite{pei_geom-gcn_2020} aim to improve the feature propagation scheme within layers of the model. More recently, researchers are proposing to make GNN models deeper. However, deeper models suffer from oversmoothing, where after stacking many GNN layers, features of the node become indistinguishable from each other, and there is a drop in the performance of the model. DropEdge \cite{rong_dropedge_2020} proposes to drop a certain number of edges to reduce the speed of convergence of oversmoothing and relieves the information loss. GCNII \cite{chen_simple_2020} use residual connections and identity mapping in GNN layers to enable deeper networks. 

\section{Experiments}
\label{experiments}
In this section, we evaluate the empirical performance of our proposed model on real-world datasets on the node classification task
and compare with other graph neural network models.

\subsection{Datasets}

For fully-supervised node classification tasks, we perform experiments on nine datasets commonly used in graph neural networks literature. Details of the datasets are presented in Table \ref{tab:fully_supervised_data}. Homophily ratio \cite{zhu_beyond_2020} denotes the fraction of edges which connects two nodes of the same label. A higher value (closer to 1) indicates strong homophily, while a lower value (closer to 0) indicates strong heterophily in the dataset. Cora, Citeseer, and Pubmed \cite{sen_collective_2008} are citation networks based datasets and in general, are considered as homophily datasets. Graphs in Wisconsin, Cornell, Texas \cite{pei_geom-gcn_2020} represent links between webpages, Actor \cite{tang_social_2009} represent actor co-occurrence in Wikipedia pages, Chameleon and Squirrel \cite{rozemberczki_multi-scale_2020} represent the web pages in Wikipedia discussing corresponding topics. These datasets are considered as heterophily datasets. To provide a fair comparison, we use publicly available data splits taken from \cite{pei_geom-gcn_2020}\footnote{https://github.com/graphdml-uiuc-jlu/geom-gcn}. These splits have been frequently used by researchers for experiments in their publications. Results of comparison methods presented in this paper are also based on this split.

In the analysis section, to demonstrate the scalability of the model for large graphs, we use ogbn-papers100M dataset\footnote{https://ogb.stanford.edu/docs/nodeprop/}, which is the largest publicly available node classification dataset. Many nodes in this dataset do not have labels assigned, hence homophily ratio is not calculated. We use standard split provided \cite{hu_open_2021} to train and evaluate the model.

\begin{table}
\centering
\caption{Statistics of the node classification datasets}
\label{tab:fully_supervised_data}
\resizebox{\linewidth}{!}{%
\begin{tabular}{lcrrrc} 
\hline
\multicolumn{1}{c}{ \textbf{Datasets} } & \textbf{Hom. Ratio} & \textbf{Nodes}  & \textbf{Edges}  & \multicolumn{1}{c}{\textbf{Features} } & \textbf{Classes}   \\ 
\hline
Cora                                    & 0.81                & 2,708           & 5,429           & 1,433                                  & 7                  \\
Citeseer                                & 0.74                & 3,327           & 4,732           & 3,703                                  & 6                  \\
Pubmed                                  & 0.80                 & 19,717          & 44,338          & 500                                    & 3                  \\
Chameleon                               & 0.23                & 2,277           & 36,101          & 2,325                                  & 4                  \\
Wisconsin                               & 0.21                & 251             & 499             & 1,703                                  & 5                  \\
Texas                                   & 0.11                & 183             & 309             & 1,703                                  & 5                  \\
Cornell                                 & 0.30                & 183             & 295             & 1,703                                  & 5                  \\
Squirrel                                & 0.22                & 5,201           & 198,353         & 2,089                                  & 5                  \\
Actor                                   & 0.22                & 7,600           & 26,659          & 932                                    & 5                  \\ 
\cmidrule(r){1-6}
\multicolumn{2}{l}{ogbn-papers100M}                           & 111,059,956     & 1,615,685,872   & 128                                    & 172                \\
\hline
\end{tabular}
}
\end{table}

\subsection{Preprocessing}

We follow the same preprocessing steps used by \cite{pei_geom-gcn_2020} and \cite{chen_simple_2020}. Other models also follow the same set of procedures. Initial node features are row-normalized. To account for both homophily and heterophily, we use the adjacency matrix and adjacency matrix with added-self loops for feature transformation. Both matrices are symmetrically normalized. For efficient computation, adjacency matrices are stored and used as sparse matrices.



\begin{table*}
\centering
\caption{Mean classification accuracy on fully-supervised node classification task. Results for GCN, GAT, GraphSAGE, Cheby+JK, MixHop and H2GCN-1 are taken from \cite{zhu_beyond_2020}. For GEOM-GCN and GCNII results are taken from the respective article. Best performance for each dataset is marked as bold and second best performance is underlined for comparison. }
\label{tab:full_super_results}
\resizebox{\linewidth}{!}{%
\begin{tabular}{lccccccccc} 
\toprule
                     & \textbf{Cora}          & \textbf{Citeseer}      & \textbf{Pubmed}        & \textbf{Chameleon}      & \textbf{Wisconsin}      & \textbf{Texas}          & \textbf{Cornell}        & \textbf{Squirrel}       & \textbf{Actor}           \\ 
\hline
\textbf{GCN}         & 87.28$\pm$1.26         & 76.68$\pm$1.64         & 87.38$\pm$0.66         & 59.82$\pm$2.58          & 59.80$\pm$6.99          & 59.46$\pm$5.25          & 57.03$\pm$4.67          & 36.89$\pm$1.34          & 30.26$\pm$0.79           \\
\textbf{GAT}         & 82.68$\pm$1.80         & 75.46$\pm$1.72         & 84.68$\pm$0.44         & 54.69$\pm$1.95          & 55.29$\pm$8.71          & 58.38$\pm$4.45          & 58.92$\pm$3.32          & 30.62$\pm$2.11          & 26.28$\pm$1.73           \\
\textbf{GraphSAGE}   & 86.90$\pm$1.04         & 76.04$\pm$1.30         & 88.45$\pm$0.50         & 58.73$\pm$1.68          & 81.18$\pm$5.56          & 82.43$\pm$6.14          & 75.95$\pm$5.01          & 41.61$\pm$0.74          & 34.23$\pm$0.99           \\
\textbf{Cheby+JK}    & 85.49$\pm$1.27         & 74.98$\pm$1.18         & 89.07$\pm$0.30         & 63.79$\pm$2.27          & 82.55$\pm$4.57          & 78.38$\pm$6.37          & 74.59$\pm$7.87          & 45.03$\pm$1.73          & 35.14$\pm$1.37           \\
\textbf{MixHop}      & 87.61$\pm$0.85         & 76.26$\pm$1.33         & 85.31$\pm$0.61         & 60.50$\pm$2.53          & 75.88$\pm$4.90          & 77.84$\pm$7.73          & 73.51$\pm$6.34          & 43.80$\pm$1.48          & 32.22$\pm$2.34           \\
\textbf{GEOM-GCN}    & 85.27                  & \textbf{77.99}         & 90.05                  & 60.90                   & 64.12                   & 67.57                   & 60.81                   & 38.14                   & 31.63                    \\
\textbf{GCNII}       & \textbf{88.01$\pm$1.33}         & 77.13$\pm$1.38                  & \textbf{90.30$\pm$0.37}         & 62.48$\pm$2.74                   & 81.57$\pm$4.98                   & 77.84$\pm$5.64                   & 76.49$\pm$4.37                   & N/A                     & N/A                      \\
\textbf{H2GCN-1}     & 86.92$\pm$1.37         & 77.07$\pm$1.64         & 89.40$\pm$0.34         & 57.11$\pm$1.58          & 86.67$\pm$4.69          & \uline{84.86$\pm$6.77}  & 82.16$\pm$4.80          & 36.42$\pm$1.89          & \textbf{35.86$\pm$1.03}  \\ 
\hline
\textbf{Ours(3-hop)} & 87.73$\pm$1.36         & 77.19$\pm$1.35         & 89.73$\pm$0.39         & \uline{78.14$\pm$1.25}  & \textbf{88.43$\pm$3.22} & \textbf{87.30$\pm$5.55} & \uline{87.03$\pm$5.77}  & \uline{73.48$\pm$2.13}  & 35.67$\pm$0.69           \\
\textbf{Ours(8-hop)} & \uline{87.93$\pm$1.00} & \uline{77.40$\pm$1.93} & \uline{89.75$\pm$0.39} & \textbf{78.27$\pm$1.28} & \uline{87.84$\pm$3.37}  & \textbf{87.30$\pm$5.28} & \textbf{87.84$\pm$6.19} & \textbf{74.10$\pm$1.89} & \uline{35.75$\pm$0.96}   \\
\bottomrule
\end{tabular}
}
\end{table*}

\subsection{Settings and Baselines}

For a fully-supervised node classification task, each dataset is split evenly for each class into 60\%, 20\%, and 20\% for training, validation, and testing. We report the performance as mean classification accuracy over 10 random splits.

We fix the embedding size to 64 and set the initial learnable scalar parameter with respect to each hop to 1 and $\gamma$ is set to 1. Thus, the initial scalar value $\alpha_i$ is set to $1/L$. Hyper-parameter settings of the model for best performance are found by performing a grid-search over a range of hyper-parameters. 

We compare our model to 8 different baselines and use the published results as the best performance of these models. GCNII \cite{chen_simple_2020} and H2GCN \cite{zhu_beyond_2020} have proposed multiple variants of their model. We have chosen the variant with the best performance on most datasets. 

\subsection{Results}
Table \ref{tab:full_super_results} shows the comparison of the mean classification accuracy of our model with other popular GNN models. On heterophily datasets, our model shows significant improvements especially 64\% on Squirrel and 23\% on Chameleon dataset. Similarly, on Wisconsin, Texas, and Cornell, improvements are 2\%, 3\%, and 7\%, respectively. H2GCN has closer performance to our model than other GNN models as its architecture design accounts for the heterophily present in class labels and distinguishes node's self-features from neighbor's features. However, with our proposed model, we are able to achieve higher accuracy. The performance of other GNN models is quite a bit lower as their design is more suitable for homophily datasets.

On homophily datasets, we observe most of the models have comparable performance with GCNII and GEOM-GCN in the lead. Our model is still comparable to state of the art and coming as second-best among various comparison measures.

\section{Discussion}
\label{discussion}

\begin{table*}
\centering
\caption{Ablation study over 1080 different hyperparameter settings.}
\label{tab:ablation_study}
\resizebox{\linewidth}{!}{%
\begin{tabular}{lccccccccc} 
\toprule
                              & \textbf{Cora}           & \textbf{Citeseer}       & \textbf{Pubmed}         & \textbf{Chameleon}      & \textbf{Wisconsin}      & \textbf{Texas}          & \textbf{Cornell}        & \textbf{Squirrel}       & \textbf{Actor}           \\ 
\hline
\textbf{Proposed}             & 83.68$\pm$2.22          & 74.48$\pm$1.44          & \textbf{89.24$\pm$0.27} & \textbf{72.48$\pm$4.16} & 81.48$\pm$5.62          & \textbf{78.80$\pm$5.88} & \textbf{78.09$\pm$2.22} & \textbf{63.57$\pm$6.83} & 33.54$\pm$1.21           \\
\textbf{Without soft-selection} & \textbf{87.07$\pm$0.26} & \textbf{76.45$\pm$0.27} & 89.09$\pm$0.39          & 72.27$\pm$1.34          & 78.03$\pm$6.55          & 76.28$\pm$6.72          & 74.32$\pm$6.54          & 61.73$\pm$4.15          & 34.15$\pm$0.64           \\
\textbf{Common weight ($W^{(0)}$)}        & 83.19$\pm$1.41          & 72.15$\pm$1.02          & 88.96$\pm$0.28          & 68.24$\pm$6.03          & 70.56$\pm$10.94         & 68.45$\pm$7.65          & 68.18$\pm$9.13          & 56.63$\pm$8.54          & 32.73$\pm$1.48           \\
\textbf{Without Hop-normalization}        & 77.12$\pm$3.49          & 71.40$\pm$10.01         & 87.72$\pm$0.77          & 53.06$\pm$6.18          & \textbf{82.60$\pm$2.68} & 76.33$\pm$3.87          & 76.18$\pm$3.43          & 32.60$\pm$6.38          & \textbf{36.66$\pm$0.55}  \\
\bottomrule
\end{tabular}
}
\end{table*}

\subsection{Ablation Studies}

In this section, we consider the effect of various proposed design strategies on the performance of the model. In general, graph neural networks are sensitive to the hyperparameters used in training and require some amount of tuning to get the best performance. Since each dataset may have different set of best hyperparameters, it can be difficult to judge design decisions based just on best performance of the model with single hyperparameter setting. To provide a comprehensive evaluation, we compare the average accuracy of the model over 1080 combinations of the hyperparameters. The hyperparameters we tune are learning rate and weight decay of layers and dropout value applied as regularization between layers.  Table \ref{tab:ablation_study} shows the average of classification accuracy values under various settings.

For most datasets, our proposed design schemes lead to better average accuracy. Cora and Citeseer show better average performance without softmax regularization, however, the peak performance is marginally less with regularization. Even though Wisconsin shows higher average accuracy without normalization, however, the best performance on the dataset was achieved with the normalization layer. We found that Actor was the only dataset where performance reduced with the addition of the normalization layer. Without the normalization layer, our model achieves 37.63\% accuracy. However, to maintain consistency, we do not include it in the main results. These variations also highlight the fact that a single set of design choices may not apply to all datasets/tasks and some level of exploration is required.

It is interesting to note that performance on almost all datasets is sensitive to the choice of the hyperparameters for training the model as there is a wide gap between best and average performance. One exception is Pubmed, where the model's performance is relatively unperturbed under various hyperparameter combinations. 

\begin{figure}[h]
    \centering
    \includegraphics[width=\linewidth]{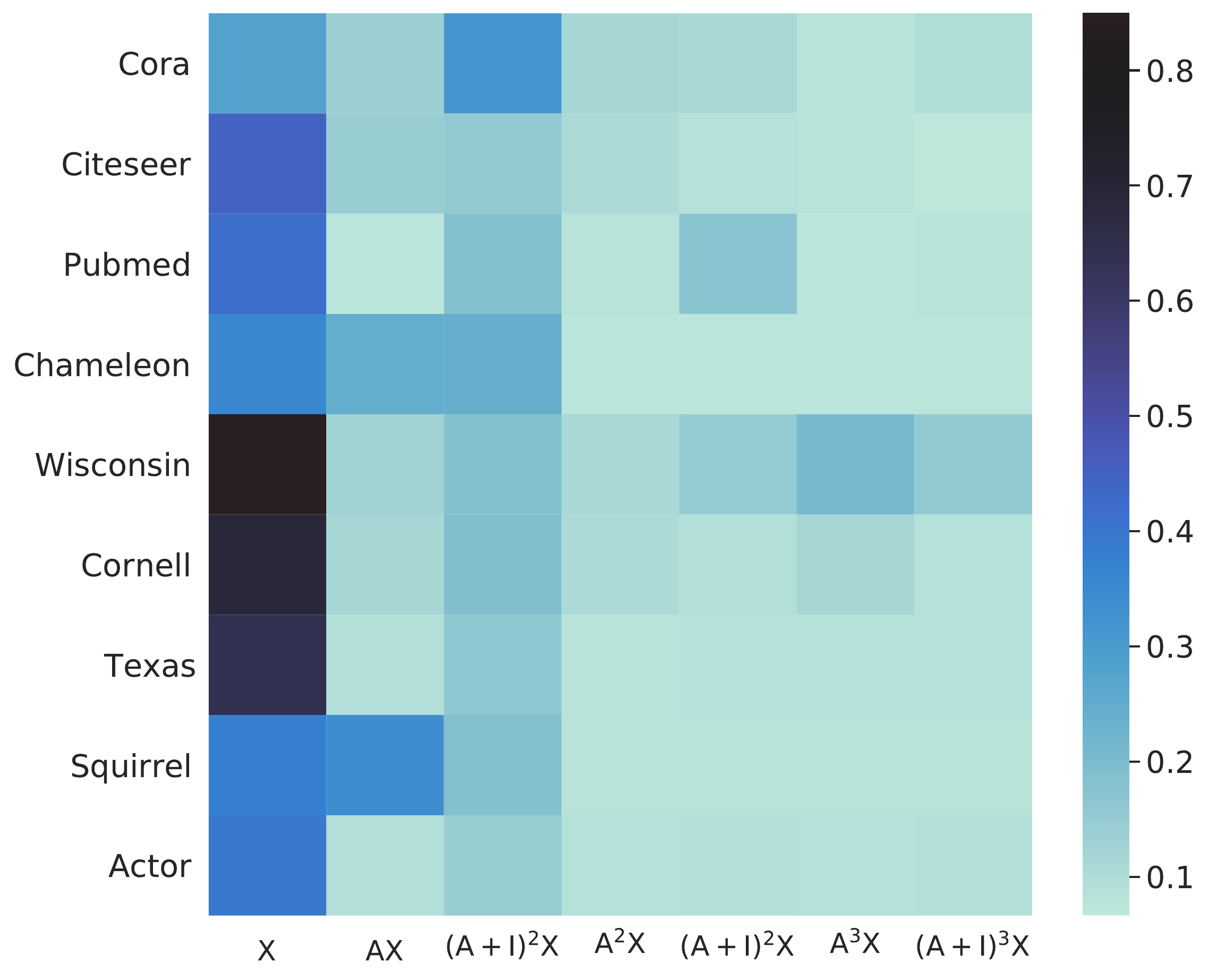}
    \caption{Heatmap of average of learned soft-selection scalar for all datasets}
    \label{fig:scalar_val_heat}
\end{figure}

\begin{figure*}
    \centering
    \includegraphics[width=0.9\textwidth]{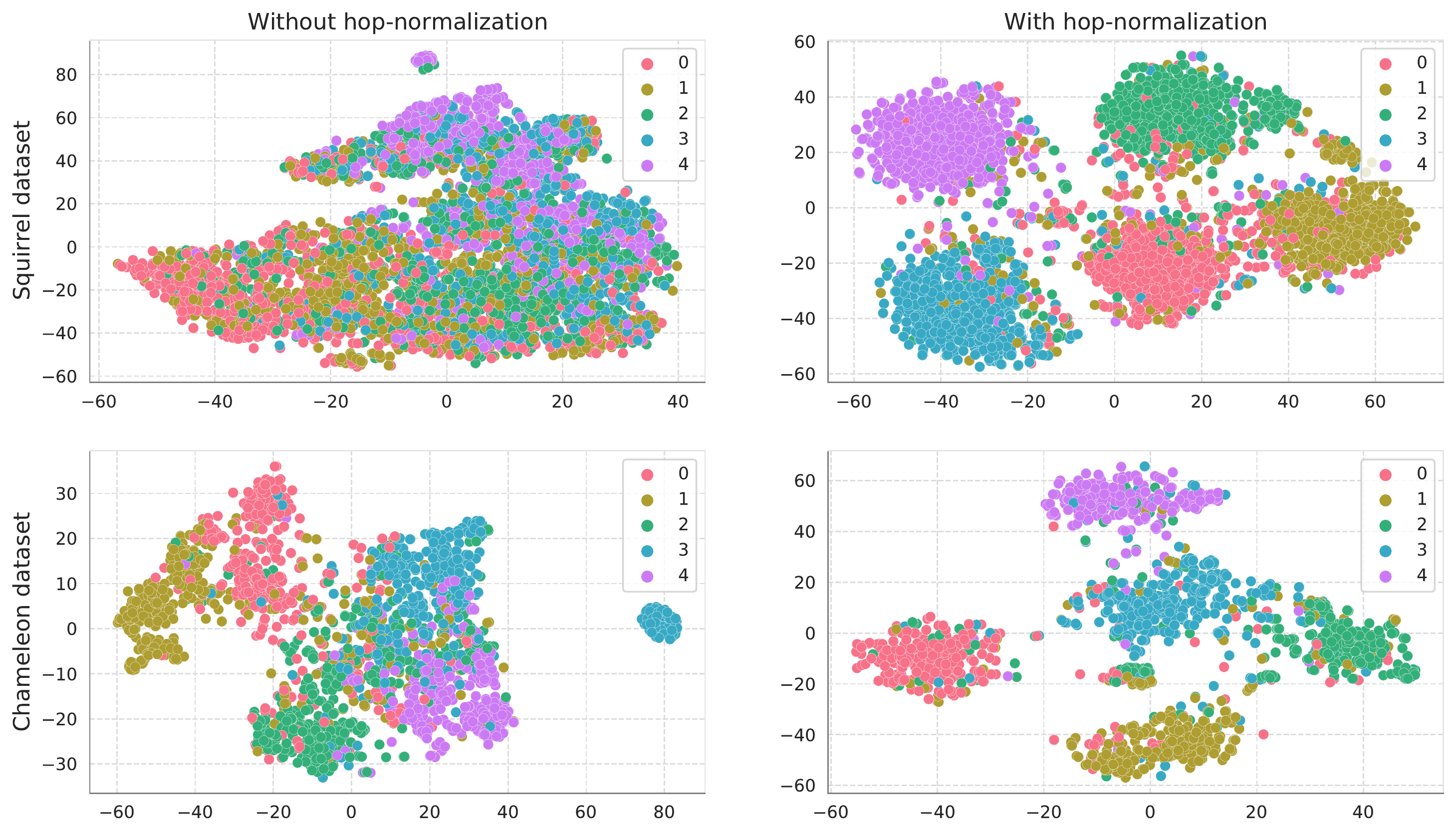}
    \caption{Figure shows t-SNE plots of trained embeddings (3-hop) of Squirrel and Chameleon datasets without (left) and with hop-normalization (right). Points represent nodes and colors represent their respective labels. Mean classification accuracy without and with hop-normalization are 39.92\% and 73.48\% for Squirrel; 61.38\% and 78.14\% for Chameleon datasets respectively.    }
    \label{fig:hop_norm}
\end{figure*}

\subsection{Soft-Selection Parameter Analysis}

We analyze the learned soft-selection parameters on average over different model hyperparameter combinations. We use four different settings: Proposed model setting, without softmax regularization on scalar weight parameters, shared linear transformation layer on input features, and without Hop-Normalization on input feature activations.  For homophily datasets, it is easy to see that self-looped features are given more importance. Among heterophily datasets, Wisconsin, Cornell, Texas, and Actor have the most weights on node's ego features. In these datasets, graph structure plays a limited role in the performance accuracy of the model. For Chameleon and Squirrel datasets, we observed that the node's own features and first-hop features(without self-loop) were more useful for classification than any other features.

\subsection{Hop-Normalization}

In our experimental results, we find Chameleon and Squirrel datasets have significant improvements. To understand the results better, we create 2-dimensional plot of the trained embeddings of both datasets using t-SNE\cite{maaten_accelerating_2014}. Figure \ref{fig:accuracy_hop} shows the comparison of embeddings with and without hop-normalization. Without hop-normalization, embeddings of the nodes are not separated clearly, thus resulting in lower classification performance. We observe similar performance on other GNN models. While with hop-normalization, the node embeddings are well separated into clusters corresponding to their label leading to a higher observed performance with our model.

\subsection{Model Scalability}

Many GNN models by design are not scalable for large graph datasets with millions of nodes. 
To demonstrate the scalability of our model, we run experiments on \texttt{ogbn-papers100M} dataset \cite{wang_microsoft_2020}\cite{hu_open_2021} which is a citation graph with about 111 million nodes, 1.6 billion edges and 172 node label classes. Similar to our previous experimental settings, we generate a set of features with $A$ and $\Tilde{A}$ for 3-hop aggregation. The dimension of the hidden layer is set to 256 and $\gamma$ is set to L=7 (equal to number of input features) to provide stable training. The model is trained batchwise with input features for 10 random initializations, and we report mean accuracy.

We compare the accuracy of our model with SGC \cite{wu_simplifying_2019},  Node2Vec \cite{grover_node2vec_2016} and SIGN \cite{frasca_sign_2020}. Similar to our method, input features can be precomputed in SGC and SIGN, thus making them scalable for larger datasets. Once features are computed, the model can be trained with small input batches of node features on the GPU. Many other GNN models cannot be trained for larger graphs as the feature generation, and model training are combined. 

Table \ref{tab:scalability_result} shows the mean node classification accuracy along with published results of other methods taken from \cite{frasca_sign_2020}\cite{hu_open_2021}. Our model outperforms all other methods, with SIGN having a closer performance to ours. However, SIGN uses the adjacency matrix of both directed and undirected versions of the graph for feature transformations, while our model only utilizes the adjacency matrix of the undirected graph.

\begin{table}[h]
\centering
\caption{Mean classification accuracy on ogbn-100M dataset. SGC result is taken from \cite{hu_open_2021} and Node2Vec and SIGN results are taken from \cite{frasca_sign_2020}. Best performance is marked bold and second best performance is underlined.}
\label{tab:scalability_result}
\begin{tabular}{ll} 
\toprule
\multicolumn{1}{c}{\textbf{Method}} & \multicolumn{1}{c}{\textbf{Accuracy}}  \\ 
\hline
\textbf{SGC}                        & 63.29$\pm$0.19                                        \\
\textbf{Node2Vec}                   & 58.07$\pm$0.28                                        \\
\textbf{SIGN}                       & \uline{65.11$\pm$0.14}                                \\
\textbf{FSGNN}                      & \textbf{67.17$\pm$0.14}                               \\
\bottomrule
\end{tabular}
\end{table}

\subsection{ Effect of increase in hops }

\begin{figure}[h]
    \centering
    \includegraphics[width=0.82\linewidth]{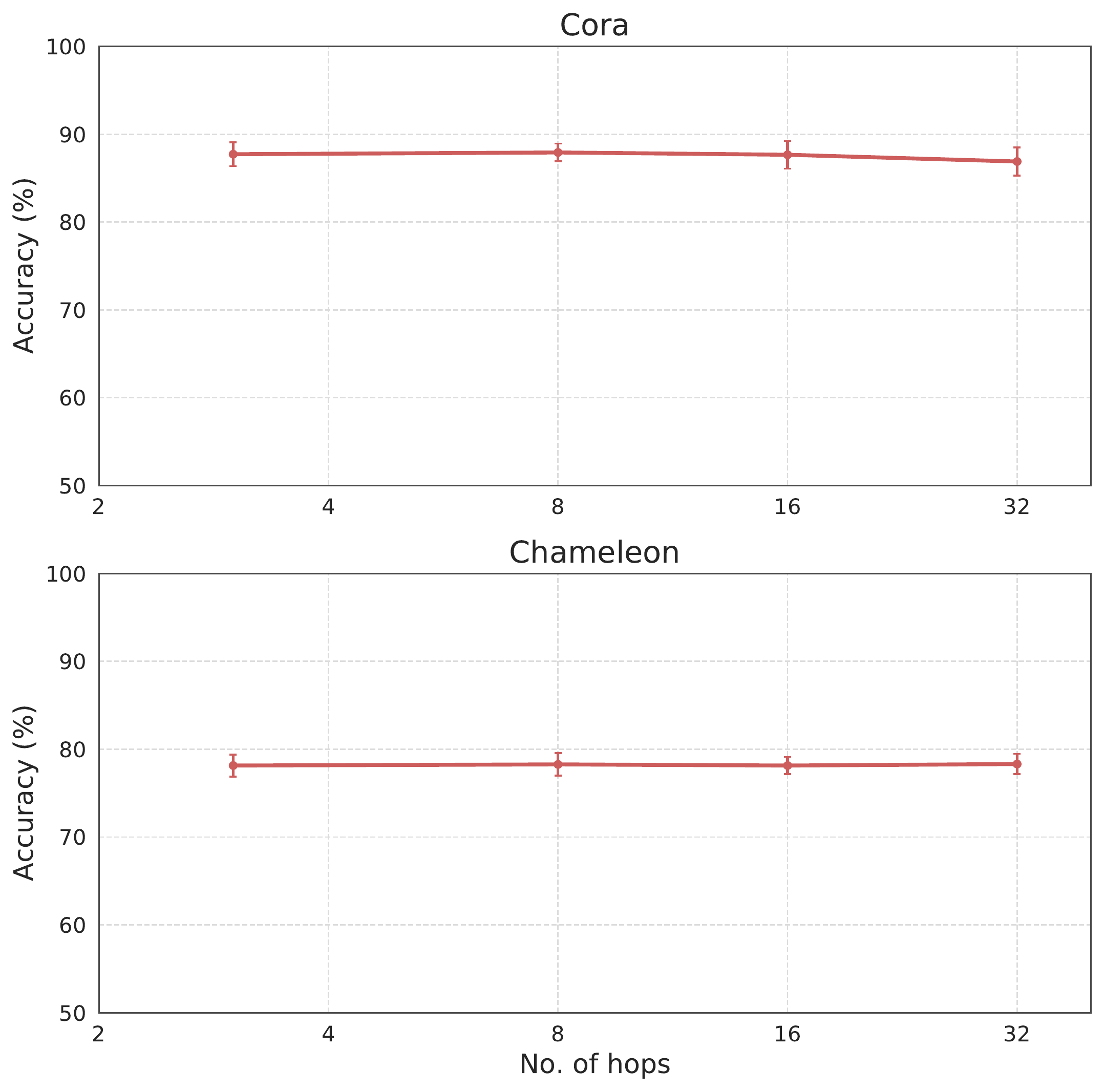}
    \caption{Figure shows the effect on classification accuracy of FSGNN with increase in the number of hops of feature aggregation on Cora (homophily) and Chameleon (heterophily) dataset. x-axis is in logarithmic scale. }
    \label{fig:accuracy_hop}
\end{figure}

In this section, we evaluate the change in model's performance with increase in the hops for aggregation. We choose one homophily dataset (Cora) and one heterophily dataset (Chameleon). Experiments are run with hop values set to 3,8,16, and 32. Figure \ref{fig:accuracy_hop} shows the performance of the model for each hop setting. We observe that there is little variation in the performance of the model. This result is intuitive as aggregated features from higher hops are not very useful, and the model can learn to place low weights on them.

\section{Conclusion}
\label{conclusion}

We discuss three GNN design strategies: separation of feature aggregation and representation learning; soft-selection of features, and hop-normalization. Using these simple and effective strategies, we propose a novel GNN model, called FSGNN. Using extensive experiments, we show that FSGNN outperforms the current state of the art GNN models on the node classification task. Analysis of the learned parameters provides us the crucial information of feature importance. Furthermore, we show that our model can be scaled for graphs with millions of nodes and billions of edges.

\vspace{5mm}
\section*{Implementation Details}

For reproducibility of experimental results, we provide the details of our experiment setup and hyperparameters of the model. 

We use PyTorch 1.6.0 as deep learning framework on Python 3.8. Model training is done on Nvidia V100 GPU with 16 GB graphics memory and CUDA version 10.2.89.

For node classfication results (\ref{tab:full_super_results}), we do grid search for learning rate and weight decay of the layers and dropout between the layers. Hyperparameters are set for first layer $fc1$, second layer $fc2$ and scalar weight parameter $sca$. ReLU is used as non-linear activation and Adam is used as the optimizer. Table \ref{tab:param_search} shows details of hyperparameter search space. Table \ref{tab:3_hop_param} and \ref{tab:8_hop_param} show the best hyperparameters for the model in 3-hop and 8-hop configuration respectively.

For experiments on ogbn-papers100M dataset, we did not do grid search. Based on the data from earlier experiments we manually tuned the hyperparameters to get the accuracy result. Batch size of 10000 was used for training data. Table \ref{tab:ogbn_papers} shows the relevant hyperparameters for the model.

\begin{table}[h]
\centering
\caption{Hyperparameter search space}
\label{tab:param_search}
\begin{tabular}{ll} 
\toprule
\textbf{Hyperparameter} & \multicolumn{1}{c}{\textbf{Values}}  \\ 
\hline
\textbf{$WD_{sca}$}          & 0.0, 0.0001, 0.001, 0.01, 0.1        \\
\textbf{$LR_{sca}$}          & 0.04, 0.02, 0.01, 0.005              \\
\textbf{$WD_{fc1}$}          & 0.0, 0.0001, 0.001                   \\
\textbf{$WD_{fc2}$}          & 0.0, 0.0001, 0.001                   \\
\textbf{$LR_{fc}$}           & 0.01, 0.005                          \\
\textbf{$Dropout$}        & 0.5, 0.6, 0.7                        \\
\bottomrule
\end{tabular}
\end{table}

\begin{table}[h]
\centering
\caption{Hyperparameters of the 3-hop model}
\label{tab:3_hop_param}
\resizebox{\linewidth}{!}{%
\begin{tabular}{lcccccc} 
\toprule
\multicolumn{1}{l}{\textbf{Datasets}} & \textbf{$WD_{sca}$} & \textbf{$LR_{sca}$} & \textbf{$WD_{fc1}$} & \textbf{$WD_{fc2}$} & \textbf{$LR_{fc}$} & \textbf{$Dropout$}  \\ 
\hline
\textbf{Cora}                         & 0.1            & 0.01           & 0.001          & 0.0001         & 0.01          & 0.6               \\
\textbf{Citeseer}                     & 0.0001         & 0.005          & 0.001          & 0.0            & 0.01          & 0.5               \\
\textbf{Pubmed}                       & 0.01           & 0.005          & 0.0001         & 0.0001         & 0.01          & 0.7               \\
\textbf{Chameleon}                    & 0.1            & 0.005          & 0.0            & 0.0            & 0.005         & 0.5               \\
\textbf{Wisconsin}                    & 0.0001         & 0.01           & 0.001          & 0.0001         & 0.01          & 0.5               \\
\textbf{Texas}                        & 0.001          & 0.01           & 0.001          & 0.0            & 0.01          & 0.7               \\
\textbf{Cornell}                      & 0.0            & 0.01           & 0.001          & 0.001          & 0.01          & 0.5               \\
\textbf{Squirrel}                     & 0.1            & 0.04           & 0.0            & 0.001          & 0.01          & 0.7               \\
\textbf{Actor}                        & 0.0            & 0.04           & 0.001          & 0.0001         & 0.01          & 0.7               \\
\bottomrule
\end{tabular}
}
\end{table}

\begin{table}[h]
\centering
\caption{Hyperparameters of the 8-hop model}
\label{tab:8_hop_param}
\resizebox{\linewidth}{!}{%
\begin{tabular}{lcccccc} 
\toprule
\multicolumn{1}{l}{ \textbf{Datasets} } & \textbf{$WD_{sca}$}  & \textbf{$LR_{sca}$}  & \textbf{$WD_{fc1}$}  & \textbf{$WD_{fc2}$}  & \textbf{$LR_{fc}$}  & \textbf{$Dropout$}   \\ 
\hline
\textbf{Cora}                           & 0.1                  & 0.02                 & 0.001                & 0.0001               & 0.01                & 0.6                  \\
\textbf{Citeseer}                       & 0.0001               & 0.01                 & 0.001                & 0.0001               & 0.01                & 0.5                  \\
\textbf{Pubmed}                         & 0.01                 & 0.02                 & 0.0001               & 0.0                  & 0.005               & 0.7                  \\
\textbf{Chameleon}                      & 0.1                  & 0.01                 & 0.0                  & 0.0                  & 0.005               & 0.5                  \\
\textbf{Wisconsin}                      & 0.001                & 0.02                 & 0.001                & 0.0001               & 0.01                & 0.5                  \\
\textbf{Texas}                          & 0.01                 & 0.01                 & 0.001                & 0.0                  & 0.01                & 0.7                  \\
\textbf{Cornell}                        & 0.0                  & 0.01                 & 0.001                & 0.0001               & 0.01                & 0.5                  \\
\textbf{Squirrel}                       & 0.1                  & 0.02                 & 0.0                  & 0.0001               & 0.01                & 0.5                  \\
\textbf{Actor}                          & 0.0001               & 0.04                 & 0.001                & 0.0001               & 0.01                & 0.7                  \\
\bottomrule
\end{tabular}
}
\end{table}


\begin{table}[h!]
\centering
\caption{Hyperparameters for the ogbn-paper100M dataset}
\label{tab:ogbn_papers}
\resizebox{\linewidth}{!}{%
\begin{tabular}{lccccccc} 
\toprule
 \textbf{Dataset}                                                   & \textbf{$WD_{sca}$}  & \textbf{$LR_{sca}$}  & \textbf{$WD_{fc1}$}  & \textbf{$WD_{fc2}$}  & \textbf{$LR_{fc1}$}& \textbf{$LR_{fc2}$}  & \textbf{$Dropout$}   \\ 
\hline
\begin{tabular}[c]{@{}l@{}}\textbf{ogbn-papers100M}\\ \end{tabular} & 0.1                  & 0.0001               & 0.001                & 0.000001             & 0.00005             & 0.0002              & 0.5                  \\
\bottomrule
\end{tabular}
}
\end{table}

\section*{Acknowledgement}

This work was supported by JSPS Grant-in-Aid for Scientific Research (Grant Number 21K12042, 17H01785), JST CREST (Grant Number JPMJCR1687), and the New Energy and Industrial Technology Development Organization (Grant Number JPNP20006)

\printbibliography

\appendix

\end{document}